\documentclass[11pt,a4paper]{article}
\usepackage{times,latexsym}
\usepackage{url}
\usepackage[T1]{fontenc}
\usepackage{latexsym}
\usepackage{lscape}
\usepackage{acl2021}
\usepackage{float}
\usepackage{multicol}
\usepackage{multirow}
\usepackage{amssymb}% http://ctan.org/pkg/amssymb
\usepackage{pifont}% http://ctan.org/pkg/pifont

\usepackage{graphicx}
\usepackage{subfigure}
\usepackage{color,colortbl}
\usepackage{xcolor}
\usepackage{microtype}
\definecolor{LightGreen}{HTML}{CCFFCC}
\definecolor{LightYellow}{HTML}{FFF5CC}
\definecolor{LightRed}{HTML}{FFB5B2}
\definecolor{LightBlue}{HTML}{CAEEF7}
\definecolor{Gray}{gray}{0.85}
\newcolumntype{a}{>{\columncolor{Gray}}c}
\newcolumntype{b}{>{\columncolor{white}}c} 
\aclfinalcopy
\usepackage{xspace,mfirstuc,tabulary}

\newif\iftaclinstructions
\taclinstructionsfalse % AUTHORS: do NOT set this to true
\iftaclinstructions
\renewcommand{\confidential}{}
\renewcommand{\anonsubtext}{(No author info supplied here, for consistency with
TACL-submission anonymization requirements)}
\newcommand{\instr}
\fi

\title{Do Encoder Representations of Generative Dialogue Models Encode Sufficient Information about the Task ?}

\author{Prasanna Parthasarathi \textsuperscript{1,2},
  Joelle Pineau \textsuperscript{1,2,4},
  Sarath Chandar \textsuperscript{2,3,4} \\
  \textsuperscript{1} School of Computer Science, McGill University \\
  \textsuperscript{2} Quebec Artificial Intelligence Institute (Mila), Canada \\
  \textsuperscript{3} \'Ecole Polytechnique de Montr\'eal \\
  \textsuperscript{4} Canada CIFAR AI Chair
  \\
}

\usepackage{microtype}
\date{}

\begin{document}
\maketitle

\begin{abstract}
Predicting the next utterance in dialogue is contingent on encoding of users' input text to generate appropriate and relevant response in data-driven approaches. Although the semantic and syntactic quality of the language generated is evaluated, more often than not, the encoded representation of input is not evaluated. As the representation of the encoder is essential for predicting the appropriate response, evaluation of encoder representation is a challenging yet important problem. In this work, we showcase evaluating the text generated through human or automatic metrics is not sufficient to appropriately evaluate soundness of the language understanding of dialogue models and, to that end, propose a set of probe tasks to evaluate encoder representation of different language encoders commonly used in dialogue models. From experiments, we observe that some of the probe tasks are easier and some are harder for even sophisticated model architectures to learn. And, through experiments we observe that RNN based architectures have lower performance on automatic metrics on text generation than transformer model but perform better than the transformer model on the probe tasks indicating that RNNs might preserve task information better than the Transformers. 
\end{abstract}

\section{Introduction}

The task of dialogue modeling requires learning through interaction, often, from humans. The model is expected to understand the input text for it to interact, and the interaction can be meaningful only when the language understanding gets better. Approaches for solving dialogue task include information retrieval based approaches like selecting a response from a set of canned responses \protect\cite{lowe2015incorporating} or keeping track of very specific information which are \emph{a priori} marked as informative slot-value pairs \protect\cite{guo2018dialog,asri2017frames} or generating the next response with token-by-token \protect\cite{vinyals2015neural,lowe2015incorporating,serban2015hierarchical, li2016deep,li2017adversarial,parthasarathi2018extending}. The evaluation of the different approaches have mostly relied on the output of the model -- the slot predicted, response selected or generated. 

The issues in evaluation -- automatic evaluation metrics uncorrelated with human judgement -- showcased by \protect\citet{liu2016not} is still an open problem. Attempts to mimic human scores for better evaluation metric \protect\cite{lowe2017towards} and other metrics that aim to correlate with the human judgement \protect\cite{sinha2020learning,tao2018ruber} evaluate the quality of the text generated but do not evaluate the language understanding component of a model. The language understanding component of an agent more often than not goes unnoticed with only token-level evaluation metrics on the generated text.

To that end, we propose evaluating the encoder representation of dialogue models through probe tasks\footnote{\href{https://github.com/ppartha03/Dialogue-Probe-Tasks-Public}{https://github.com/ppartha03/Dialogue-Probe-Tasks-Public}} constructed from the commonly used dialogue data sets -- MultiWoZ \protect\cite{budzianowski2018multiwoz} and PersonaChat \protect\cite{zhang2018personalizing}. Concretely, we use the representation learnt by the encoders while training on dialogue generation tasks to solve a set of dialogue related classification tasks as a proxy to probe the information encoded in the encoder representation.
%Evaluating model's understanding through probe tasks \protect\cite{belinkov2019analysis,jawahar2019does,anand2019unsupervised} is of recent interest.
%A probe task is a backward reasoning task, where a model reasons out its understanding of the input through simple questions as classification task.
We study the performance of language encoders in 17 different probe tasks with varying degree of difficulties -- binary classification, multi-label classification and multi-label prediction. For example, predicting whether the current dialogue has single or multiple tasks, identifying the number of tasks, identifying the tasks, presence of a specific information provided by the user among many others. The probe tasks allow a way to quantify the understanding of a model and help identify biases, if any, in the task of dialogue prediction. We observed the performance of the models in the probe tasks to little fluctuate with different seed values thus allowing to analyse the encoder representation with minimal variance. Further, the experiments on probe tasks help in understanding deeper differences in between recurrent neural network (RNN) and Transformer encoders that were previously not evident from the token-level evaluation methods.

Our contributions in the paper are:
\begin{itemize}
    \item Showcasing the significantly high variance in human evaluation of dialogues.
    \item Proposing a list of probe tasks -- 2 semantic, 13 information specific and 3 downstream as an alternate evaluation of dialogue systems.
    \item Finding that the representation learnt by recurrent neural network based models is better at solving the probe tasks than the ones based on transformer model.
\end{itemize}
% \section{Background}
% A dialogue task is modeled as a conditional language generation task where the model takes in a sequence of words (context), generates as output a sequence of tokens (response). A sequence modeling approach to solve these tasks is by an encoder-decoder architecture where the encoder and decoder are any choice of RNN, LSTM \protect\cite{lstm}, GRU \protect\cite{cho2014gru}, or the Bi-Directional versions of them. Recently, transductive style approach as in Transformer \protect\cite{vaswani2017attention} is also used for similar tasks. Some approaches have also tried to directly model the user intent and thereby learn an intent conditional language generation \protect\cite{serban2017deep}. For the scope of the experiments in this paper, we use Transformer and 4 SEQ2SEQ models -- BiLSTM with Attention \protect\cite{bidirectional}, Seq2Seq \protect\cite{vinyals2015neural}, Seq2Seq with Attention \protect\cite{nmtbahadanau} and HRED \protect\cite{serban2015hierarchical}.
\section{Related Work}

Evaluating dialogue models has been an important topic of study. While many of the metrics have focussed on evaluating the generated text through n-gram overlap based heuristics -- BLEU \cite{papineni2002bleu}, ROUGE \cite{lin2004rouge}, METEOR \cite{lavie2007meteor} -- there have also been learned metrics like ADEM \cite{lowe2017towards}, MAudE \cite{sinha2020learning}, RUBER \cite{tao2018ruber} among other metrics \cite{celikyilmaz2020evaluation}. Though language generation has been an important component of study, there are not many studies that benchmark soundness of encoding information by dialogue systems.

Probe tasks in language generation \protect\cite{conneau2018you,belinkov2019analysis,elazar2020bert} has been used to understand the information encoded in continuous embedding of sentences. Such probe tasks are set up as classification tasks that are solved with model learnt representation. As it is easier to control the biases in probe tasks than in the downstream tasks, research in language generation has analysed models on probe tasks like using encoder representation to identify words in input (\textbf{WordCont}) to measuring encoder sensitivity to shifts in bigrams \protect\cite{conneau2018you,belinkov2019analysis}.

Analysis using probe tasks has been done also in reinforcement learning (RL).  \protect\citet{anand2019unsupervised} learn state representation for an RL agent in an unsupervised setting and introduce a set of probe tasks to evaluate the representation learnt by agents. This includes using an annotated data set with markers for position of the agent, current score, items in inventory, target's location among others. The authors train a shallow linear classifier to identify specific entities in the embedded input that serves as a metric for the representational soundness of the learning algorithm.

Applications of computer vision like caption generation for images \protect\cite{vinyals2015show} or videos \protect\cite{donahue2015long} use attention based models to parse over the hidden states of a convolutional neural network (ConvNet) \protect\cite{lecun1998gradient}. The attention over the ConvNet features are visualized to observe the words corresponding to different parts of the image. Visualizing the attention has been one of the qualitative probe task for text generation conditioned on images \protect\cite{xu2015show}.

\section{Dialogue probe Tasks}

\begin{table*}[!t]
    \centering
    \footnotesize
    \begin{tabular}{|p{1.5cm}|l|p{7cm}|p{1cm}|p{2cm}|}
     \hline
     \textbf{Task} & \textbf{Task Name} & \textbf{Description} & \textbf{\#Classes} & \textbf{Multi-Label Prediction} \\
     \hline
      \multirow{2}{1cm}{Semantic}&\cellcolor{LightYellow}\textbf{UtteranceLoc}$^{*}$&\cellcolor{LightYellow} How long has the conversation been happening ? & \cellcolor{LightYellow}5 & \cellcolor{LightYellow}No\\
      &\cellcolor{LightYellow}\textbf{WordCont}$^{+}$ & \cellcolor{LightYellow}Which mid-frequency word is encoded in the context ? & \cellcolor{LightYellow}1000 & \cellcolor{LightYellow}No\\
      \hline
       \multirow{13}{1cm}{Information Specific}&\cellcolor{LightGreen}\textbf{IsMultiTopic} & \cellcolor{LightGreen}Does the conversation have more than one topic ? & \cellcolor{LightGreen} 2 & \cellcolor{LightGreen} No\\
       &\cellcolor{LightGreen}\textbf{NumAllTopics} &\cellcolor{LightGreen} How many topics does this conversation have ? & \cellcolor{LightGreen} 6 & \cellcolor{LightGreen}No\\
       &\cellcolor{LightGreen}\textbf{RepeatInfo} & \cellcolor{LightGreen}Which information provided by the user is repeated ? & \cellcolor{LightGreen}11 & \cellcolor{LightGreen}Yes \\
       &\cellcolor{LightGreen}\textbf{NumRepeatInfo} & \cellcolor{LightGreen}What many number of recent information are repeats ? & \cellcolor{LightGreen}7 & \cellcolor{LightGreen}No \\
       &\cellcolor{LightYellow}\textbf{AllTopics}  & \cellcolor{LightYellow}What are all the topics discussed so far ? & \cellcolor{LightYellow}8 & \cellcolor{LightYellow}No\\
       &\cellcolor{LightYellow}\textbf{RecentSlots} &\cellcolor{LightYellow} What is the \textit{recent} information given by the user ? & \cellcolor{LightYellow} 37 & \cellcolor{LightYellow} Yes\\
       &\cellcolor{LightYellow}\textbf{NumRecentInfo} &\cellcolor{LightYellow} How \emph{many} information did the user provide \textit{recently} ?& \cellcolor{LightYellow}10 & \cellcolor{LightYellow}No\\
       &\cellcolor{LightRed}\textbf{RecentValues} &\cellcolor{LightRed} What are the details of the \textit{recent} information ? & \cellcolor{LightRed}1060 & \cellcolor{LightRed}Yes\\
       &\cellcolor{LightRed}\textbf{AllSlots} &\cellcolor{LightRed} What \textit{all} information are given by the user so far? & \cellcolor{LightRed}37& \cellcolor{LightRed}Yes\\
       &\cellcolor{LightRed}\textbf{AllValues} & \cellcolor{LightRed}What are the details of \textit{all} the information provided ? & \cellcolor{LightRed}1060 &\cellcolor{LightRed}Yes \\
       &\cellcolor{LightRed}\textbf{RecentTopic} &\cellcolor{LightRed} What is the current topic of the dialogue ? & \cellcolor{LightRed}8&\cellcolor{LightRed}No\\
       &\cellcolor{LightRed}\textbf{NumAllInfo} &\cellcolor{LightRed}How \emph{many} information did the user provide so far ? & \cellcolor{LightRed} 20& \cellcolor{LightRed}No\\
       &\cellcolor{LightRed}\textbf{PersonalInfo}$^{+}$ &\cellcolor{LightRed}What keywords in USER persona does the model identify? &\cellcolor{LightRed} 3754& \cellcolor{LightRed}Yes\\
     \hline
     \multirow{3}{1cm}{Downstream task}&\cellcolor{LightRed}\textbf{ActionSelect}&\cellcolor{LightRed} Which downstream task (database query) follows the current conversation ? & \cellcolor{LightRed} 32& \cellcolor{LightRed} No \\
       &\cellcolor{LightRed}\textbf{EntitySlots} &\cellcolor{LightRed}What information is required to construct the query ? & \cellcolor{LightRed}29 & \cellcolor{LightRed}Yes \\
       &\cellcolor{LightRed}\textbf{EntityValues} &\cellcolor{LightRed}What values should be passed to the query ? & \cellcolor{LightRed} 1309 & \cellcolor{LightRed} Yes\\
       \hline
       
    \end{tabular}
    \caption{The difficulty levels of different tasks is measured with the average performance of an untrained encoder. There is a natural grading in the selection of tasks that expects better language understanding to solve. $^+$ indicate the task is present both in MultiWoZ and PersonaChat datasets. 
    $^*$ indicate the task is only in PersonaChat. If no indicator is present, the task is evaluated only in MultiWoZ dataset.}
    \label{tab:probe-tasks-list}
\end{table*}

Like other tasks, dialogue task requires a learning agent to have sufficient understanding of the context to generate a response; at times the models have been shown to not have basic understanding leading to incorrect response prediction.  Although dialogue models are evaluated on grammar, semantics, and relevance of the generated text, seldom has that been extended to evaluate the language encoding capacity of these models. The tasks proposed and discussed in this paper are shown in Table \ref{tab:probe-tasks-list}. 

\subsection{Basic Probe Tasks} The basic probe tasks evaluate if the encoder representation can be used to predict the existence of a mid-frequency token in the context (\textit{WordCont}) \protect\cite{belinkov2019analysis}, or test if the encoding of the context provides information of how long the dialogue has been going on (\textit{UtteranceLoc}) \protect\cite{sinha2020learning}. For \textit{UtteranceLoc} task,  the conversation is split into 5 different temporal blocks and a classifier trained on the encoded context embedding is used to predict the appropriate label.

\subsection{Information Specific Probe Tasks} 

We construct 12 information specific probe tasks to evaluate if specific information is retained in the encoder representation of input text. The information specific tasks have different levels of difficulty. For example, \emph{IsMultiTopic} is a binary classification task, \emph{NumAllTopics} is a multi-label classification task while \emph{AllTopics} is a multi-label prediction.

\subsection{Downstream probe Tasks}

Further we evaluate the language understanding of dialogue models on their performance on relevant downstream tasks. Towards evaluating the model's understanding of the user utterance, the downstream probe tasks verify if the encoder representation allows to predict the user dialogue act. The dialogue state tracking measures the performance of a model on such tasks \protect\cite{henderson2014second} but seldom is it evaluated on generative dialogue models. \protect\citet{neelakantan2019neural} use entity, values and action information to train on the dialogue generation task but the performance of a generative dialogue model without explicitly training on the downstream tasks are not compared. Towards that, we propose \textbf{ActionSelect}, \textbf{EntitySlots}, \textbf{EntityValues} probe tasks. The details of the task are shown in Table \ref{tab:probe-tasks-list}.

\section{Experiments}
\subsection{Data sets}
With the probe tasks we study different dialogue encoder architectures trained on next utterance generation on MultiWoZ 2.0 \protect\cite{budzianowski2018multiwoz} and PersonaChat \protect\cite{zhang2018personalizing} data sets. The features of the data sets are shown in Table \ref{tab:dataset}.
\begin{table}[h]
    \centering
    \small
    \begin{tabular}{c|c|c|c}
    \hline
    {\bf Data set} & {\bf Train } & {\bf Validation} & {\bf Vocabulary} \\
    \hline
         {\it PersonaChat} & $\sim$ 10900 & 1500 & 16k \\
         {\it MultiWoZ} & $\sim$ 8400 & 1000 & 13k \\
    \hline
    \end{tabular}
    \caption{Distribution of the dialogues in the data sets.}
    \label{tab:dataset}
\end{table}
To comprehensively compare several model selection criteria, we experimented with selecting models based on BLEU \protect\cite{papineni2002bleu}, ROUGE-F1 \cite{lin2004rouge}, METEOR \cite{lavie2007meteor} and Vector-Based (Average BERT embedding) metrics. We present the results from BLEU as a selection criteria in the paper. Further in the Appendix we compare the evolution of the performance of different models in the probe tasks over the entire training.

The classification tasks for probing the encoder representation are constructed for every generated response that requires information from the dialogue history thus far. We split the probe tasks in Train/Test/Valid corresponding to the splits the tasks are constructed from. First, we train the dialogue models on end-to-end dialogue generation and use the encoder representation to train and test on the probe tasks. To that, we store the encoder parameters after every epoch during dialogue generation training and compute the results of probe tasks after every epoch.

\subsection{Models}
We train 5 commonly used encoder architectures on the task of next utterance generation in the two data sets.

\paragraph{\textsc{LSTM Encoder-Decoder}} The architecture \protect\cite{vinyals2015neural} has an LSTM cell to encode the input context only in the forward direction. For a sequence of words in the input context $(w_1^i, w_2^i, \ldots, w_{T'}^i)$ LSTM encoder generates $\{h_t\}_1^T$. The decoder LSTM's hidden state is initialized with $h_t^T$ and the decoder outputs one token at each step of decoding. For the experiments, we used two layer LSTM cell where the first layer applies recurrent operation on the input to the model and the layer above recurs on the outputs of the layer below. The encoder final hidden state (from the 2nd layer) is passed as an input to the decoder. We train the model with cross entropy loss as shown in Equation \ref{eqn:cross-entropy}.
\begin{equation}
    \sum_{t=1}^{T} -y_t \log (p\left(\hat{y}_t)\right) - (1-y_t) \log (1- p\left(\hat{y}_t)\right) 
    \label{eqn:cross-entropy}
\end{equation}
where $y_t$ is the $t^{th}$ ground truth token distribution in the output sequence, $\hat{y}_t$ is model generated token and $p$ is the model learned distribution over the tokens.
We train the model with Adam \protect\cite{kingma2014adam} optimizer with teacher forcing \protect\cite{teacherforcing}.
\paragraph{\textsc{LSTM Encoder-Attention Decoder}} The architecture is similar to the LSTM Encoder-Decoder with an exception of an attention module to the decoder. The attention module \protect\cite{nmtbahadanau} linearly combines the encoder hidden states ${h_t}_1^{T}$ as an input to the decoder LSTM at every step of decoding, unlike only having the last encoder hidden state.

\paragraph{\textsc{Hierarchical Recurrent Encoder Decoder}} The model has encoding done by two encoder modules acting at different levels \protect\cite{sordoni2015hierarchical}; \textit{sentence encoder} to encode the sentences that feeds in as input to the \textit{context encoder}. Both the encoders are LSTMs. The decoder is an attention decoder.
\paragraph{\textsc{Bi-LSTM Encoder-Attention Decoder}} The encoder is a concatenation of two LSTMs that can read the input from forward and backward direction \protect\cite{bidirectional}. The hidden state is computed as the summation of the hidden states of the two encoders. The decoding is done with an attention decoder.
\paragraph{\textsc{Transformer Architecture}} This state-of-the-art architecture \protect\cite{vaswani2017attention,rush2018annotated} is a transductive model that has multiple layers of attention to predict the output. We used the architecture in an encoder-decoder style by splitting half the layers for encoding and the remainder for decoding. We perform the probe tasks on the encoder hidden state computed as an average over word token attention.

The size of the models used in the experiments are detailed in Table \ref{tab:size-of-models} in Appendix. For the probe tasks, we select the untrained model, model with the best BLEU score on validation, and model from the last training epoch. %We truncate the input and pass the last 100 tokens to the model.
We use packages pytorch \protect\cite{paszke2017pytorch} and scikit-learn \protect\cite{scikit-learn} for our experiments.

\subsection{Motivation for Dialogue Probe Tasks}
\label{sec:motivation}
The texts generated by the models are largely dependent on the choice of seed values and a slight variation could result in a model generating a very different response. Although the automatic metrics have greater agreement on the score across seed values, we see that human participants do not agree on the consistency of the generated response. We pose and evaluate an alternate hypothesis where we expect the participants to identify two responses to be similar when selected from different runs of the same model with different seed values that have similar BLEU scores.

\begin{table}[h!]
    \centering
    \small
    \begin{tabular}{c|c|c}
    \hline
         {\bf Model}& {\bf PersonaChat} & {\bf MultiWoZ}  \\
    \hline
        \rowcolor{Gray}{\it BiLSTM + Attn} & 4.4 $\pm$ 0.06 & 15.5 $\pm$ 0.05 \\
        {\it Seq2Seq} & 4.5 $\pm$ 0.06 & 15.8 $\pm$ 0.17 \\
        \rowcolor{Gray}{\it Seq2Seq + Attn} & 4.4 $\pm$ 0.15 & 15.7 $\pm$ 0.11\\
        {\it HRED} & 3.9 $\pm$ 0.01 & 12.2 $\pm$ 4.00 \\
        \rowcolor{Gray}{\it Transformer} & 7.9 $\pm$ 0.17 & 29.4 $\pm$ 0.61\\
    \hline
    \end{tabular}
    \caption{BLEU scores of the models from runs with different seeds on PersonaChat and MultiWoZ data set. (Higher the better. We measure BLEU-2 (case insensitive).}
    \label{tab:bleu-scores}
\end{table}

\begin{figure}[h!]
    \centering
    \includegraphics[width=0.8\columnwidth]{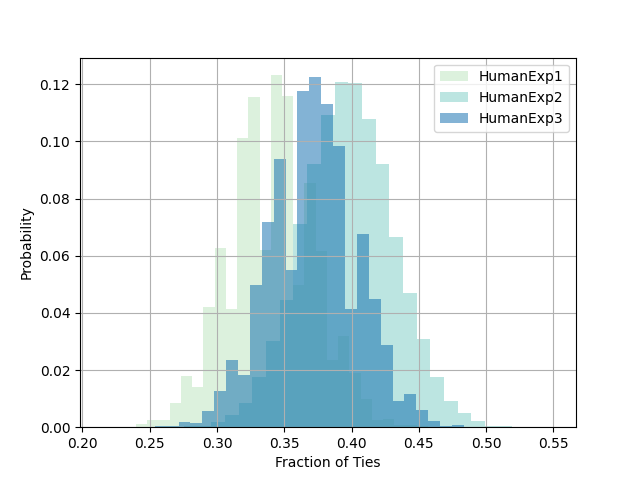}
    \caption{The mean of the distribution of tie in three different experiments was centered around 35\%, showing that the subjective scores on responses by humans are not sufficient to evaluate a model. }
     \label{fig:human_bias_goal}
\end{figure}

For the study, we sample 2000 context-response pairs in MultiWoZ dataset from the model with lower variance in BLEU score (Table \ref{tab:bleu-scores}) -- Bi-LSTM Attention -- from its two different runs. We ask the participants to select the response that is more \emph{relevant} to the given context, similar to \protect\citet{li2015diversity}. The annotators can select either of the responses or a Tie\footnote{The human evaluation proposal was evaluated and approved by an IRB.}. For every context-response pair, we collect 3 feedback from different participants (Distribution corresponding to the 3 different human responses are shown with legend HumanExp1, HumanExp2 and HumanExp3 in Figure \ref{fig:human_bias_goal}). 

Usually human evaluation is done on 100-500 responses. To understand the variance in this set up and the lack of information at the token generation level, we sample 50000 sets of 200 human responses from the collected 2000 responses and compute the fraction of times there was a tie. We observed that distribution over the fraction of times the human participants selected a \textit{Tie} was centered around 35\% (Figure \ref{fig:human_bias_goal}) with all of the probability mass within 50\%. This shows that (a) text generated by the same model produce significantly different responses with different seed values (b) attributing the choice of seed value to the performance of a model creates confusion in the evaluation because the two seeds had similar BLEU scores. The results show that evaluating only based on the text generated by a model is not suggestive of the information encoding capacity of the encoder representation. Also, the dependence of the model generated text on seed value raises a valid concern; whether a model parameter initialized with a specific seed value mimic the token generation of a model that actually encodes sufficient information in the context. The lack of clarity leads to inconclusiveness of studies with human evaluation to show whether the dialogue models have sufficient information encoded to solve the task effectively.

\subsection{Probe Tasks}
We train the models with the two dialogue data sets on next utterance generation. To understand the evolution on the probe task, we compare with 3 different parameter configurations of every model -- \emph{Untrained}, \emph{Last epoch}, and \emph{BestBLEU}. We use Logistic Regression classifier\footnote{Also, we trained a nonlinear model --multi-layer perceptron for probe tasks and the results are similar. The discussion in the paper is agnostic to the choice of the classifier.} implementation from scikit-learn \protect\cite{scikit-learn} with default parameters except the max\_iter set to 250 for all the probe tasks. The evaluation metric is \emph{F1}-score with micro averaging in multi-class prediction tasks.

\paragraph{\textsc{Probe Tasks on PersonaChat}} The models are evaluated on three probe tasks (Table \ref{tab:persona-probe-tasks}) -- two basic and one information specific. {\it UtteranceLoc} and {\it WordCont} measures if the encoded context suggests semantic awareness of the model while \textit{PersonalInfo} measures the amount of knowledge the model has about its persona from encoding of conversation history. In other words, it evaluates the extent to which persona can be identified from the context encoding with a linear classifier. A better performance in these tasks indicate that the context encoding preserves information on persona and the temporal order of the dialogue.

The PersonalInfo task is not very specific to identifying personal information but acts as an indicator to the information embedded in dialogues that goes unnoticed in the encoding. It was surprising to see that no model scored a reasonable \textit{F1}. Although Transformer model scored higher on BLEU, (Table \ref{tab:bleu-scores}) the performance of transformer on PersonalInfo task was decreasing throughout the training epochs(Table \ref{tab:persona-probe-tasks}%, also in other tasks as observed in Figures \ref{fig:probetasks_table1_personachat}, \ref{fig:probetasks_table1_multiwoz}, and \ref{fig:probetasks_table2_multiwoz} in Appendix
).
\begin{table}[h!]
\scriptsize
\centering
      \begin{tabular}{c|a|b|a}
         \hline
         \multicolumn{4}{c}{PersonaChat data set}\\
         \hline
         {\bf Model} & {\bf UtteranceLoc} & {\bf WordCont} & {\bf PersonalInfo} \\
         \hline
         \rowcolor{LightBlue}\multicolumn{4}{c}{Bi-LSTM Seq2Seq + Attention} \\
         \hline
         Untrained &37.0 $\pm$ 0.1 &43.5 $\pm$ 0.0 &0.0 $\pm$ 0.0    \\
         LastEpoch & 56.5 $\pm$ 0.0 &39.9 $\pm$ 0.0 &0.0 $\pm$ 0.0   \\
         BestBLEU & 57.2 $\pm$ 0.1 &39.7 $\pm$ 0.1 &0.0 $\pm$ 0.0  \\
        \hline
         \rowcolor{LightBlue}\multicolumn{4}{c}{HRED - LSTM} \\
         \hline
         Untrained & 1.2 $\pm$ 0.0 &51.7 $\pm$ 0.0 &0.0 $\pm$ 0.0 \\
         LastEpoch &12.8 $\pm$ 4.9 &49.4 $\pm$ 0.3 &0.0 $\pm$ 0.0   \\
         BestBLEU & 10.8 $\pm$ 3.5 &51.0 $\pm$ 0.1 &0.0 $\pm$ 0.0  \\
         \hline
         \rowcolor{LightBlue}\multicolumn{4}{c}{LSTM Seq2Seq + Attention} \\
         \hline
         Untrained & 39.9 $\pm$ 0.0 &47.2 $\pm$ 0.1 &0.0 $\pm$ 0.0 \\
         LastEpoch & 52.0 $\pm$ 0.0 &40.0 $\pm$ 0.0 &0.0 $\pm$ 0.0  \\
         BestBLEU & 54.1 $\pm$ 0.16 &43.8 $\pm$ 0.2 &0.0 $\pm$ 0.0  \\
         \hline
         \rowcolor{LightBlue}\multicolumn{4}{c}{LSTM Seq2Seq} \\
         \hline
         Untrained & 40.2 $\pm$ 0.0 &46.9 $\pm$ 0.0 &0.0 $\pm$ 0.0   \\
         LastEpoch & 50.9 $\pm$ 0.1 &40.0 $\pm$ 0.0 &0.0 $\pm$ 0.0    \\
         BestBLEU & 52.2 $\pm$ 0.1 &40.2 $\pm$ 0.0 &0.0 $\pm$ 0.0  \\
         \hline
         \rowcolor{LightBlue}\multicolumn{4}{c}{Transformer Architecture} \\
         \hline
         Untrained & 53.0 $\pm$ 0.0 &35.9 $\pm$ 0.0 &2.4 $\pm$ 0.0  \\
         LastEpoch & 42.7 $\pm$ 0.1 &46.9 $\pm$ 0.1 &0.0 $\pm$ 0.0   \\
         BestBLEU & 40.7 $\pm$ 0.1 &46.2 $\pm$ 0.0 &0.0 $\pm$ 0.0   \\
         \hline
    \end{tabular}
    \caption{Performance of different models on the probe tasks on PersonaChat data set. The performance is measured as \emph{F-1} score (Higher the better).}
    \label{tab:persona-probe-tasks}
\end{table}

The tasks UtteranceLoc and WordCont evaluate if encoder representations are indicative of how far in the conversation is the model in and identify mid-frequency words in the target response respectively. Bi-LSTM model performed the best in UtteranceLoc while the Transformer model was not in the top 3. 

We observe that the inductive biases of the RNN-based models enable random projections that are informative even without training. This correlates with independent observations on the results in \protect\cite{worldmodelblog} that argues random projections of temporal information hold non-negligible information. Similar observations are also made from the untrained Transformer model's performance on the probe tasks. 

\begin{figure*}[h!]
\centering
\subfigure[Seq2Seq Model after Last Epoch]{
    \includegraphics[width=0.95\columnwidth]{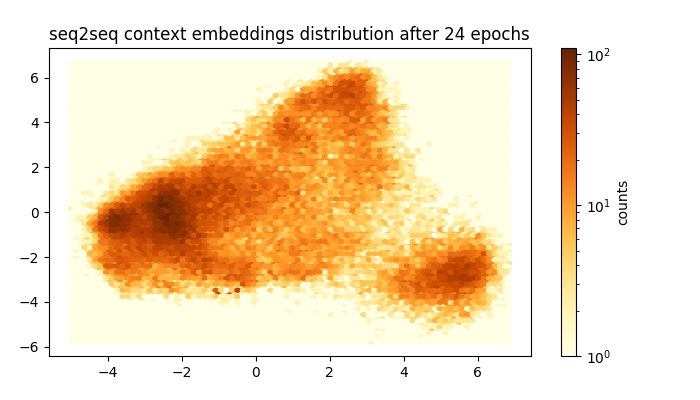}
}
\subfigure[Seq2Seq Attention Model after Last Epoch]{
    \includegraphics[width=0.95\columnwidth]{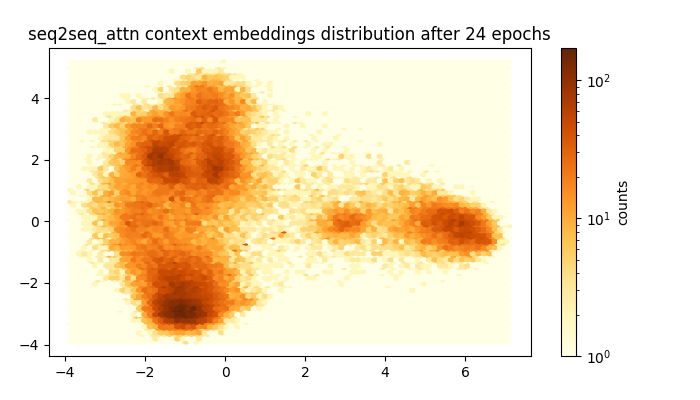}
}
\\
\subfigure[Bi-LSTM Attention Model after Last Epoch]{
    \includegraphics[width=0.95\columnwidth]{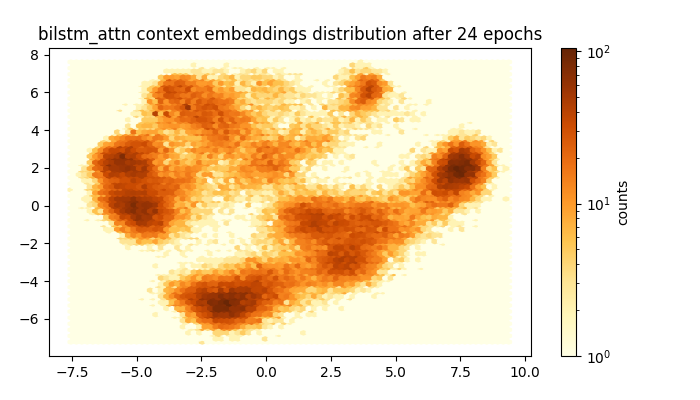}
}
\subfigure[Transformer Model after Last Epoch]{
    \includegraphics[width=0.95\columnwidth]{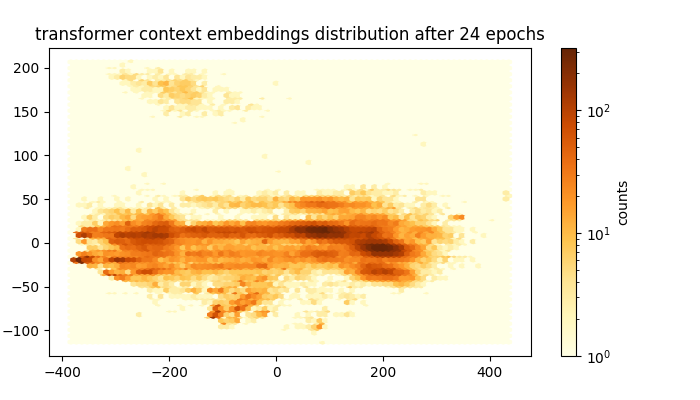}
}
\caption{Downsampled encoder hidden states on MultiWoZ data set with PCA show that Transformer model has high capacity to encode a large data set unlike the SEQ2SEQ models.}
\label{fig:multi-woz-context}
\end{figure*}

The RNN encoders project the context to a smaller manifold with its recurrent multiplication that regularizes its representation to observe structures, whereas Transformer network's attention operations project the context on to a larger manifold that prevents loss in encoding \footnote{\protect\citet{ramsauer2020hopfield} showed recently that the transformer model is a large look-up table. Our empirical results support the authors' view.} making the representation useful for the end task (Figure \ref{fig:multi-woz-context} ). This explains the RNN based encoders performing well on UtteranceLoc while Transformer model performing well on WordCont. The difference between the two classes of models is much more evident on the probe tasks in MultiWoZ data set.

\paragraph{\textsc{Probe Tasks on MultiWoZ}} 

In majority of information specific tasks and in the downstream tasks (Table \ref{tab:multiwoz-probe-tasks}), we observed that RNN based models perform significantly better than the Transformer model. Interestingly, we observed a pattern in Transformer in the two data sets, that the model's performance on the probe tasks decreased from the beginning of training till the end on all of the tasks, while for the rest of the models there was learning involved. %(Appendix \ref{fig:probetasks_table1_multiwoz}, \ref{fig:probetasks_table2_multiwoz}, and \ref{fig:probetasks_table1_personachat}). 
  \begin{table*}[h!]
        \scriptsize
        \centering
        \subfigure[]{\begin{tabular}{c|a|b|a|b|a|b|a|b}
         \hline
         \multicolumn{9}{c}{MultiWoZ data set}\\
         \hline
         {\bf Model} & {\bf UtteranceLoc} & {\bf RecentTopic} & {\bf RecentSlots} & {\bf RecentValues} & {\bf RepeatInfo} & {\bf NumRepeat}& {\bf NumRecent} & {\bf AllSlots} \\
         \hline
         \rowcolor{LightBlue}\multicolumn{9}{c}{LSTM Seq2Seq + Attention} \\
         \hline
         Untrained &46.5 $\pm$ 0.5 &35.3 $\pm$ 0.0 &39.1 $\pm$ 0.0 &30.8 $\pm$ 0.0 &64.2 $\pm$ 0.0 &69.0 $\pm$ 0.0 &41.4 $\pm$ 0.0 &30.2 $\pm$ 0.0 \\
         LastEpoch &56.5 $\pm$ 0.1 &87.1 $\pm$ 0.0 &65.6 $\pm$ 0.0 &42.2 $\pm$ 0.0 &64.9 $\pm$ 0.0 &70.0 $\pm$ 0.0 &61.7 $\pm$ 0.1 &51.5 $\pm$ 0.0\\
         BestBLEU &58.0 $\pm$ 0.1 &89.0 $\pm$ 0.0 &66.5 $\pm$ 0.0 &41.1 $\pm$ 0.0 &64.5 $\pm$ 0.0 &67.0 $\pm$ 0.0 &63.4 $\pm$ 0.0 &52.0 $\pm$ 0.1  \\
        \hline
         \rowcolor{LightBlue}\multicolumn{9}{c}{HRED - LSTM} \\
         \hline
         Untrained & 45.3 $\pm$ 1.6 &32.9 $\pm$ 0.0 &41.2 $\pm$ 0.0 &31.7 $\pm$ 0.0 &71.0 $\pm$ 0.0 &74.9 $\pm$ 0.0 &40.7 $\pm$ 0.0 &19.8 $\pm$ 0.0  \\
         LastEpoch & 38.0 $\pm$ 10.9 &54.2 $\pm$ 22.6 &36.3 $\pm$ 10.1 &21.3 $\pm$ 3.4 &69.4 $\pm$ 0.1 &74.0 $\pm$ 0.0 &39.5 $\pm$ 11.7 &32.8 $\pm$ 8.4  \\
         BestBLEU & 38.7 $\pm$ 11.3 &50.1 $\pm$ 20.5 &34.3 $\pm$ 9.3 &20.4 $\pm$ 3.1 &71.0 $\pm$ 0.1 &74.5 $\pm$ 0.1 &39.3 $\pm$ 11.6 &30.3 $\pm$ 7.9 \\
         \hline
         \rowcolor{LightBlue}\multicolumn{9}{c}{LSTM Seq2Seq } \\
         \hline
         Untrained &46.6 $\pm$ 0.3 &35.9 $\pm$ 0.0 &39.7 $\pm$ 0.0 &32.0 $\pm$ 0.0 &64.8 $\pm$ 0.0 &69.2 $\pm$ 0.0 &43.0 $\pm$ 0.0 &29.5 $\pm$ 0.0 \\
         LastEpoch &55.0 $\pm$ 0.1 &87.6 $\pm$ 0.0 &66.0 $\pm$ 0.0 &41.9 $\pm$ 0.0 &66.1 $\pm$ 0.0 &69.8 $\pm$ 0.0 &61.0 $\pm$ 0.0 &51.6 $\pm$ 0.0 \\
         BestBLEU&56.3 $\pm$ 0.0 &88.6 $\pm$ 0.0 &66.9 $\pm$ 0.0 &41.6 $\pm$ 0.0 &65.9 $\pm$ 0.0 &70.2 $\pm$ 0.0 &62.6 $\pm$ 0.0 &52.6 $\pm$ 0.0  \\
         \hline
         \rowcolor{LightBlue}\multicolumn{9}{c}{Bi-LSTM Seq2Seq + Attention} \\
         \hline
         Untrained & 44.3 $\pm$ 0.0 &50.7 $\pm$ 0.1 &35.3 $\pm$ 0.0 &27.3 $\pm$ 0.0 &64.6 $\pm$ 0.0 &70.6 $\pm$ 0.0 &39.9 $\pm$ 0.0 &36.7 $\pm$ 0.0 \\
         LastEpoch & 57.2 $\pm$ 0.0 &86.7 $\pm$ 0.0 &63.3 $\pm$ 0.0 &38.2 $\pm$ 0.0 &66.6 $\pm$ 0.0 &70.8 $\pm$ 0.0 &60.2 $\pm$ 0.1 &53.4 $\pm$ 0.0   \\
         BestBLEU &57.5 $\pm$ 0.1 &89.0 $\pm$ 0.0 &64.5 $\pm$ 0.0 &39.6 $\pm$ 0.0 &68.5 $\pm$ 0.0 &72.2 $\pm$ 0.0 &62.3 $\pm$ 0.0 &56.0 $\pm$ 0.0   \\
         \hline
         \rowcolor{LightBlue}\multicolumn{9}{c}{Transformer Architecture} \\
         \hline
         Untrained &51.2 $\pm$ 0.0 &80.3 $\pm$ 0.0 &45.6 $\pm$ 0.0 &30.6 $\pm$ 0.0 &70.4 $\pm$ 0.0 &73.3 $\pm$ 0.0 &47.5 $\pm$ 0.0 &62.9 $\pm$ 0.0  \\
         LastEpoch & 33.7 $\pm$ 0.5 &32.1 $\pm$ 1.9 &26.2 $\pm$ 1.9 &22.1 $\pm$ 1.7 &70.7 $\pm$ 0.0 &74.6 $\pm$ 0.0 &33.6 $\pm$ 3.3 &21.3 $\pm$ 0.6 \\
         BestBLEU &32.0 $\pm$ 0.5 &31.7 $\pm$ 5.3 &29.6 $\pm$ 0.3 &25.3 $\pm$ 0.2 &72.2 $\pm$ 0.0 &75.9 $\pm$ 0.0 &37.8 $\pm$ 0.4 &22.8 $\pm$ 1.44 \\
         \hline
    \end{tabular}}
\\
    \subfigure[]{\begin{tabular}{c|a|b|a|b|a|b|a|b}
         \hline
         \multicolumn{9}{c}{MultiWoZ data set}\\
         \hline
         {\bf Model} & {\bf AllValues} & {\bf NumAllInfo} & {\bf AllTopics}&{\bf NumAllTopics} & {\bf IsMultiTask} & {\bf EntitySlots} & {\bf EntityValues} & {\bf ActionSelect} \\
         \hline
         \rowcolor{LightBlue}\multicolumn{9}{c}{LSTM Seq2Seq + Attention} \\
         \hline
         Untrained & 12.6 $\pm$ 0.0 &7.0 $\pm$ 0.0 &45.1 $\pm$ 0.0 &70.3 $\pm$ 0.0 &80.1 $\pm$ 0.0 &28.0 $\pm$ 0.0 &19.6 $\pm$ 0.0 &28.7 $\pm$ 0.0  \\
         LastEpoch & 19.3 $\pm$ 0.0 &29.3 $\pm$ 0.0 &73.4 $\pm$ 0.0 &76.3 $\pm$ 0.0 &81.5 $\pm$ 0.0 &43.5 $\pm$ 0.0 &28.4 $\pm$ 0.0 &56.2 $\pm$ 0.0 \\
         BestBLEU  &18.7 $\pm$ 0.0 &29.2 $\pm$ 0.0 &74.3 $\pm$ 0.0 &76.9 $\pm$ 0.1 &82.1 $\pm$ 0.0 &42.6 $\pm$ 0.0 &29.1 $\pm$ 0.0 &56.9 $\pm$ 0.0 \\
        \hline
         \rowcolor{LightBlue}\multicolumn{9}{c}{HRED - LSTM} \\
         \hline
         Untrained & 5.3 $\pm$ 0.0 &0.0 $\pm$ 0.0 &37.5 $\pm$ 0.0 &77.6 $\pm$ 0.0 &84.2 $\pm$ 0.0 &24.9 $\pm$ 0.1 &19.0 $\pm$ 0.0 &27.3 $\pm$ 0.01  \\
         LastEpoch & 8.7 $\pm$ 0.7 &19.1 $\pm$ 2.8 &48.7 $\pm$ 18.0 &69.2 $\pm$ 3.7 &73.5 $\pm$ 4.7 &27.1 $\pm$ 5.6 &20.2 $\pm$ 3.1 &38.8 $\pm$ 11.3 \\
         BestBLEU  & 8.4 $\pm$ 0.8 &18.0 $\pm$ 2.6 &46.6 $\pm$ 16.9 &68.6 $\pm$ 3.5 &73.2 $\pm$ 4.6 &24.8 $\pm$ 4.9 &20.1 $\pm$ 3.0 &34.4 $\pm$ 9.3 \\
         \hline
         \rowcolor{LightBlue}\multicolumn{9}{c}{LSTM Seq2Seq } \\
         \hline
         Untrained &13.3 $\pm$ 0.0 &6.3 $\pm$ 0.0 &43.0 $\pm$ 0.0 &73.3 $\pm$ 0.0 &80.4 $\pm$ 0.1 &27.3 $\pm$ 0.0 &20.3 $\pm$ 0.0 &29.0 $\pm$ 0.0 \\
         LastEpoch  & 19.5 $\pm$ 0.0 &28.8 $\pm$ 0.0 &72.8 $\pm$ 0.0 &75.7 $\pm$ 0.0 &81.2 $\pm$ 0.0 &44.0 $\pm$ 0.0 &30.7 $\pm$ 0.0 &56.7 $\pm$ 0.0  \\
         BestBLEU  &18.8 $\pm$ 0.00 &29.7 $\pm$ 0.02 &74.3 $\pm$ 0.03 &77.1 $\pm$ 0.0 &81.9 $\pm$ 0.0 &44.1 $\pm$ 0.0 &28.9 $\pm$ 0.03 &57.2 $\pm$ 0.0 \\
         \hline
         \rowcolor{LightBlue}\multicolumn{9}{c}{Bi-LSTM Seq2Seq + Attention} \\
         \hline
         Untrained & 14.9 $\pm$ 0.0 &10.9 $\pm$ 0.1 &56.8 $\pm$ 0.0 &71.4 $\pm$ 0.0 &79.5 $\pm$ 0.0 &24.2 $\pm$ 0.0 &19.0 $\pm$ 0.0 &26.1 $\pm$ 0.0   \\
         LastEpoch & 20.0 $\pm$ 0.0 &28.5 $\pm$ 0.0 &74.8 $\pm$ 0.0 &78.4 $\pm$ 0.0 &84.0 $\pm$ 0.0 &42.1 $\pm$ 0.0 &29.6 $\pm$ 0.0 &55.4 $\pm$ 0.0  \\
         BestBLEU & 20.0 $\pm$ 0.0 &29.6 $\pm$ 0.0 &77.4 $\pm$ 0.0 &79.1 $\pm$ 0.0 &84.2 $\pm$ 0.0 &41.6 $\pm$ 0.0 &28.2 $\pm$ 0.0 &56.5 $\pm$ 0.0  \\
         \hline
         \rowcolor{LightBlue}\multicolumn{9}{c}{Transformer Architecture} \\
         \hline
         Untrained  & 39.6 $\pm$ 0.0 &27.3 $\pm$ 0.0 &81.2 $\pm$ 0.0 &77.6 $\pm$ 0.0 &82.8 $\pm$ 0.0 &30.3 $\pm$ 0.0 &19.7 $\pm$ 0.1 &38.5 $\pm$ 0.2 \\
         LastEpoch  & 5.1 $\pm$ 0.1 &11.5 $\pm$ 0.5 &47.7 $\pm$ 1.3 &71.9 $\pm$ 0.0 &82.0 $\pm$ 0.0 &13.5 $\pm$ 0.4 &13.4 $\pm$ 0.0 &6.8 $\pm$ 0.1  \\
         BestBLEU  & 5.6 $\pm$ 0.1 &7.3 $\pm$ 0.2 &50.4 $\pm$ 1.1 &73.5 $\pm$ 0.0 &81.7 $\pm$ 0.0 &23.3 $\pm$ 0.1 &12.2 $\pm$ 0.3 &7.8 $\pm$ 0.2 \\
         \hline
    \end{tabular}}
    \caption{ F1 scores of generative dialogue models on probe tasks in MultiWoZ dialogue data set (higher the better). SEQ2SEQ models perform significantly better than Transformer model on the probe tasks, despite the models falling behind in BLEU score. The Transformer model's performance decreased from initial to last epoch in majority of the tasks while SEQ2SEQ models have a learning curve.}
    \label{tab:multiwoz-probe-tasks}
\end{table*}

The downsampled encoder representation of the encoded contexts with PCA to 2 components (Figure \ref{fig:multi-woz-context}) shows that the range of the two axes are different for RNN-based and Transformer models. The context encoding of transformers lie in a much larger manifold. The attention layers help in spreading the data in a large manifold thereby the model can retain almost all of the generation task related information it was trained on. This can be observed in higher BLEU score the model achieves in language generation. But, the reverse of generalizing from a small data is hard to come by because the model does not have sufficient direct information to cluster except the surface level signal of predicting the right tokens. This helps the Transformer model to perform well on the token prediction task in language modelling, while abstracting information and generalizing appears to be a difficult task as is observed from its performance on probe tasks.

The RNN-based models have inductive biases to squish the input through \texttt{tanh} or \texttt{sigmoid} operations. From the visualizations and from other results, we hypothesize that this aids the model in learning a regularized representation in a low-data set up. But, this can potentially be unhelpful when the input is a large set of samples and has rich structure as that requires a model to aggressively spread out. Transformer architecture can thrive in such a set up and that can be validated by the performance of large Transformer models like GPT \protect\cite{GPT}, GPT-2 \protect\cite{GPT2}, GPT-3 \protect\cite{GPT3}, BERT \protect\cite{devlin2018bert,devlin2019bert}, RoBERTa \protect\cite{liu2019roberta} etc., whereas the results in the probe tasks show that RNN-based models are adept at learning unsupervised structures for better understanding of the input. Also we note that the performance in probe tasks can be a pseudo metric to measure the capacity of the model in generalizing to unobserved structures in inputs in a low data scenario.

\section{Discussion}
\label{sec:discussion}
Systematic evaluation of language understanding through probe tasks is important to analyze the correlation between input and output in complex language understanding tasks. We observed that most of the data collected for dialogue generation tasks \cite{lowe2015ubuntu,ritter2011data} do not provide tasks to sanity check language understanding through probing encoder representations. Absence of probe tasks lead to draw imperfect correlations like the one between token-level accuracy and model's encoding of dialogue information from the context. At this point one may wonder, why not train the model with all the probe tasks as auxiliary tasks for an improved performance ? Although it is a possibility, such a set up does not evaluate a model's ability to generalize to understanding in unseen dialogues. One could potentially train a model with a fraction of the probe tasks as auxiliary tasks and evaluate on the rest, we leave that for future work.

It is also interesting to draw parallels to \emph{Unit Testing} in software engineering \protect\cite{koomen1999test}, where the smallest software components of a system are tested for their design and logical accuracy. The difference between a deterministic application software and a stochastic decision making ML module is that the behavior of the ML system is data-driven while for a software system it is driven by logic. Despite the difference, the unit testing and probe tasks could share a common ground towards ensuring the better representation of the encoded contexts.

\begin{table}[h]
    \centering
    \small
    \begin{tabular}{p{1.5cm}|p{1.3cm}|p{1.3cm}|p{1.3cm}}
    \hline
         {\bf Model}& {\bf Easy} & {\bf Medium} & {\bf Hard}  \\
    \hline
        \rowcolor{Gray}{\it LSTM-Attn} & 77.6$\pm$6.2 & 65.7$\pm$7.6 & 44.4$\pm$23.7 \\
        {\it HRED} & 72.1$\pm$2.7 & 39.3$\pm$5.1 & 25.4$\pm$13.6 \\
        \rowcolor{Gray}{\it Seq2Seq} & 77.2$\pm$5.3 & 65.7$\pm$7.6 & 44.9$\pm$23.5\\
        {\it BiLSTM} & 78.5$\pm$6.2 & 65.6$\pm$8.7 & 44.2$\pm$23.3 \\
        \rowcolor{Gray}{\it Transformer} & 77.2$\pm$4.9 & 43.3$\pm$14.7 & 24.4$\pm$16.4\\
    \hline
    \end{tabular}
    \caption{Aggregate F1 scores of the models on performance in probe tasks on MultiWoZ data set. }
    \label{tab:difficulty-graded}
\end{table}
{\textsc{Dialogue Models}} As an alternate to token-level evaluation, comparison of different model architectures can be meaningfully made with an aggregate metric on the probe tasks in three groups of difficulty -- \emph{easy} ((Ave. SEQ2SEQ) Untrained F1 $>$ .50), \emph{medium}(0.25 $<$ Untrained F1 $\geq$ 0.50), and \emph{hard} (Untrained F1 $<$ 0.25). Such an analysis, as shown in Table \ref{tab:difficulty-graded}, allows better inspection of the model's language understanding and a fairer comparison between the models. We can see from Table \ref{tab:difficulty-graded} that the models have difficulty in learning to solve hard probe tasks from the encoder representations. The results can be used to build novel inductive biases for neural architectures that address one or a group of aspects in the language understanding of dialogue prediction models. 

{\textsc{Dialogue data sets}} The challenges in dialogue modeling has been evolving majorly because of the complex data sets. But, data sets on chit-chat dialogues often have little to no auxiliary tasks to evaluate the dialogue management abilities of a model. This limits the practitioners to validate the models only on the text generation task which, in this paper, is shown to have little to no correlation with the model's ability to understanding the encoded summary of natural language context. 

\section{Conclusion}
We propose a set of probe tasks to evaluate the encoder representation of end-to-end generative dialogue models. We observed that mimicking surface level token prediction do not reveal much about a model's  ability to understand a natural language context. The results on probe tasks showed that RNN-based models perform better than transformer model in encoding information in the context. We also found some probe tasks that all of the models find difficult to solve; this invites novel architectures that can handle the language understanding aspects in dialogue generation. Although language generation is required for a dialogue model, the performance in token/response prediction alone cannot be a proxy for the model's ability to understand a conversation. Hence, systematically identifying issues in language understanding through probe tasks can help in building better models and collecting challenging data sets. 

\section*{Acknowledgements}

We would like to acknowledge Compute Canada and Calcul Quebec for providing computing resources used in this work. The authors would also like to thank members of Chandar Research Lab, Mila for helping with the code reviews and reviewing the manuscripts. Sarath Chandar and Joelle Pineau are supported by Canada CIFAR AI Chair, and Sarath Chandar is also supported by an NSERC Discovery Grant.

\bibliography{emnlp2020}
\bibliographystyle{acl_natbib}

\newpage
\appendix
\section*{Appendix}
\section{Model Parameters}
\begin{table}
    \centering
    \small
    \begin{tabular}{p{5cm}|c}
    \hline
       {\bf Model}  & {\bf Parameters}  \\
    \hline
        \rowcolor{Gray}{\it LSTM Encoder-Decoder} &	11M \\
        {\it LSTM Encoder-Decoder + Attention} &	11M \\
        \rowcolor{Gray}{\it HRED} &	12M \\
        {\it Bi-LSTM Encoder-Decoder}	& 12M\\
        \rowcolor{Gray}{\it Transformer}	& 41M \\
    \hline
    \end{tabular}
    \caption{Size of parameters of the models used in all the experiments on the two data sets. \emph{M} for Million.}
    \label{tab:size-of-models}
\end{table}

\label{sec:model-parameters}
\begin{itemize}
    \item For SEQ2SEQ models, we used a 256 unit hidden size LSTM with 2 layers and a 128 unit input embedding dimension. The learning rate we used for all the models is 4E-3. 
    \item For Transformer, we used a 512 unit hidden size, 512 unit input embedding dimension, 2 attention header and 4 layers. 
    \item We used Adam as the optimizer to optimize on the cross-entropy loss. 
    \item We averaged the results over 3 different seeds.
    \item We used a truncated history of last 100 tokens as context to keep the training uniform across the models.
\end{itemize}

\begin{table*}[h!]
        \tiny
        \centering
        \begin{tabular}{c|a|b|a|b|a|b|a|b}
         \hline
         \multicolumn{9}{c}{MultiWoZ Dataset}\\
         \hline
         {\bf Model} & {\bf UtteranceLoc} & {\bf RecentTopic} & {\bf RecentSlots} & {\bf RecentValues} & {\bf RepeatInfo} & {\bf NumRepeatInfo}& {\bf NumRecentInfo} & {\bf AllSlots} \\
         \hline
         \rowcolor{LightBlue}\multicolumn{9}{c}{LSTM Seq2Seq + Attention} \\
         \hline
         BERT &37.12 $\pm$ 2.59 &42.74 $\pm$ 16.78 &43.53 $\pm$ 4.54 &30.93 $\pm$ 0.63 &70.82 $\pm$ 0.01 &74.71 $\pm$ 0.01 &44.76 $\pm$ 1.89 &23.94 $\pm$ 6.31 \\
         F1 &58.08 $\pm$ 0.09 &89.31 $\pm$ 0.08 &66.72 $\pm$ 0.02 &39.55 $\pm$ 0.05 &71.25 $\pm$ 0.01 &75.10 $\pm$ 0.00 &62.14 $\pm$ 0.02 &52.57 $\pm$ 0.10 \\
         BLEU & 57.55 $\pm$ 0.05 &89.91 $\pm$ 0.07 &67.39 $\pm$ 0.02 &40.49 $\pm$ 0.04 &70.92 $\pm$ 0.00 &74.73 $\pm$ 0.00 &62.48 $\pm$ 0.02 &53.08 $\pm$ 0.11 \\
         METEOR &58.23 $\pm$ 0.08 &89.26 $\pm$ 0.08 &66.83 $\pm$ 0.02 &39.72 $\pm$ 0.04 &71.29 $\pm$ 0.00 &75.01 $\pm$ 0.00 &62.23 $\pm$ 0.01 &52.58 $\pm$ 0.10\\
        \hline
         \rowcolor{LightBlue}\multicolumn{9}{c}{HRED - LSTM} \\
         \hline
         BERT &18.78 $\pm$ 10.58 &23.78 $\pm$ 16.97 &16.41 $\pm$ 8.07 &10.44 $\pm$ 3.27 &71.78 $\pm$ 0.02 &75.51 $\pm$ 0.01 &19.27 $\pm$ 11.14 &13.31 $\pm$ 5.31 \\
         F1 &37.18 $\pm$ 10.38 &49.59 $\pm$ 19.55 &33.95 $\pm$ 8.98 &20.81 $\pm$ 3.26 &71.33 $\pm$ 0.01 &74.99 $\pm$ 0.01 &38.49 $\pm$ 11.14 &28.49 $\pm$ 6.63 \\
         BLEU & 37.15 $\pm$ 10.35 &50.98 $\pm$ 20.94 &34.84 $\pm$ 9.69 &20.63 $\pm$ 3.21 &71.68 $\pm$ 0.00 &75.06 $\pm$ 0.00 &38.59 $\pm$ 11.18 &30.23 $\pm$ 7.84 \\
         METEOR &41.04 $\pm$ 5.85 &50.78 $\pm$ 20.86 &44.50 $\pm$ 2.49 &28.96 $\pm$ 0.18 &71.72 $\pm$ 0.00 &75.28 $\pm$ 0.00 &50.71 $\pm$ 1.44 &30.21 $\pm$ 7.84 \\
         \hline
         \rowcolor{LightBlue}\multicolumn{9}{c}{LSTM Seq2Seq } \\
         \hline
         BERT &54.16 $\pm$ 0.94 &63.24 $\pm$ 16.20 &55.13 $\pm$ 4.78 &34.62 $\pm$ 0.76 &72.00 $\pm$ 0.00 &75.90 $\pm$ 0.00 &54.06 $\pm$ 2.46 &37.48 $\pm$ 5.21  \\
         F1 &57.56 $\pm$ 0.06 &89.44 $\pm$ 0.04 &68.00 $\pm$ 0.00 &40.98 $\pm$ 0.03 &71.22 $\pm$ 0.01 &75.32 $\pm$ 0.01 &62.78 $\pm$ 0.01 &53.07 $\pm$ 0.04\\
         BLEU& 57.37 $\pm$ 0.06 &89.45 $\pm$ 0.03 &68.08 $\pm$ 0.01 &39.78 $\pm$ 0.07 &71.28 $\pm$ 0.01 &75.36 $\pm$ 0.01 &62.33 $\pm$ 0.05 &53.40 $\pm$ 0.05 \\
         METEOR & 57.84 $\pm$ 0.04 &89.03 $\pm$ 0.01 &67.74 $\pm$ 0.01 &40.37 $\pm$ 0.10 &71.10 $\pm$ 0.00 &74.75 $\pm$ 0.00 &61.85 $\pm$ 0.00 &53.04 $\pm$ 0.02 \\
         \hline
         \rowcolor{LightBlue}\multicolumn{9}{c}{Bi-LSTM Seq2Seq + Attention} \\
         \hline
         BERT & 57.98 $\pm$ 0.03 &78.79 $\pm$ 3.19 &57.24 $\pm$ 1.71 &35.59 $\pm$ 0.34 &71.35 $\pm$ 0.00 &75.18 $\pm$ 0.01 &57.57 $\pm$ 0.18 &48.37 $\pm$ 1.11\\
         F1 &57.99 $\pm$ 0.05 &89.63 $\pm$ 0.03 &64.85 $\pm$ 0.00 &39.16 $\pm$ 0.00 &71.76 $\pm$ 0.01 &75.30 $\pm$ 0.01 &60.85 $\pm$ 0.07 &54.68 $\pm$ 0.04  \\
         BLEU & 59.04 $\pm$ 0.10 &89.85 $\pm$ 0.03 &65.03 $\pm$ 0.00 &39.06 $\pm$ 0.00 &71.98 $\pm$ 0.01 &75.63 $\pm$ 0.00 &60.36 $\pm$ 0.05 &54.96 $\pm$ 0.05  \\
         METEOR &58.45 $\pm$ 0.07 &89.28 $\pm$ 0.02 &64.21 $\pm$ 0.00 &39.19 $\pm$ 0.00 &71.54 $\pm$ 0.00 &75.35 $\pm$ 0.01 &60.49 $\pm$ 0.05 &54.65 $\pm$ 0.04  \\
         \hline
         \rowcolor{LightBlue}\multicolumn{9}{c}{Transformer Architecture} \\
         \hline
         BERT &39.11 $\pm$ 0.09 &58.38 $\pm$ 0.14 &29.97 $\pm$ 0.00 &24.50 $\pm$ 0.01 &72.39 $\pm$ 0.01 &76.02 $\pm$ 0.00 &38.80 $\pm$ 0.01 &43.37 $\pm$ 0.17 \\
         F1 & 39.89 $\pm$ 0.21 &67.44 $\pm$ 0.44 &33.37 $\pm$ 0.14 &24.96 $\pm$ 0.02 &72.75 $\pm$ 0.01 &76.26 $\pm$ 0.00 &40.43 $\pm$ 0.05 &51.19 $\pm$ 0.51\\
         BLEU &39.46 $\pm$ 0.00 &57.05 $\pm$ 1.50 &30.10 $\pm$ 0.27 &23.72 $\pm$ 0.03 &72.70 $\pm$ 0.00 &75.97 $\pm$ 0.00 &39.11 $\pm$ 0.08 &40.43 $\pm$ 1.21\\
         METEOR &38.50 $\pm$ 0.25 &56.26 $\pm$ 1.87 &30.98 $\pm$ 0.11 &24.94 $\pm$ 0.02 &72.26 $\pm$ 0.01 &75.79 $\pm$ 0.00 &39.47 $\pm$ 0.04 &38.70 $\pm$ 1.59\\
         \hline
    \end{tabular}
    \caption{Comparison of models selected different selection metrics on probe tasks in MultiWoZ dialogue data set. The performance is measured with \emph{F1} on the probetasks.}
    \label{tab:multiwoz-probe-tasks-1-metric}
\end{table*}

\begin{table*}[h!]
\tiny
\centering
    \begin{tabular}{c|a|b|a|b|a|b|a|b}
         \hline
         \multicolumn{9}{c}{MultiWoZ Dataset}\\
         \hline
         {\bf Metric} & {\bf AllValues} & {\bf NumAllInfo} & {\bf AllTopics}&{\bf NumAllTopics} & {\bf IsMultiTask} & {\bf EntitySlots} & {\bf EntityValues} & {\bf ActionSelect} \\
         \hline
         \rowcolor{LightBlue}\multicolumn{9}{c}{LSTM Seq2Seq + Attention} \\
         \hline
         BERT & 6.16 $\pm$ 0.34 &8.52 $\pm$ 2.18 &49.07 $\pm$ 5.13 &77.98 $\pm$ 0.00 &84.97 $\pm$ 0.01 &27.49 $\pm$ 1.30 &22.22 $\pm$ 0.47 &30.25 $\pm$ 6.74  \\
         F1 &12.54 $\pm$ 0.01 &26.54 $\pm$ 0.02 &75.22 $\pm$ 0.03 &79.56 $\pm$ 0.02 &84.70 $\pm$ 0.01 &41.74 $\pm$ 0.02 &31.20 $\pm$ 0.03 &60.00 $\pm$ 0.00\\
         BestBLEU  &12.81 $\pm$ 0.01 &25.73 $\pm$ 0.02 &75.33 $\pm$ 0.02 &79.39 $\pm$ 0.02 &85.30 $\pm$ 0.00 &41.29 $\pm$ 0.03 &31.57 $\pm$ 0.03 &60.14 $\pm$ 0.01\\
         METEOR & 12.53 $\pm$ 0.01 &26.62 $\pm$ 0.02 &75.21 $\pm$ 0.03 &79.52 $\pm$ 0.02 &84.67 $\pm$ 0.01 &41.70 $\pm$ 0.02 &31.48 $\pm$ 0.02 &60.06 $\pm$ 0.00 \\
        \hline
         \rowcolor{LightBlue}\multicolumn{9}{c}{HRED - LSTM} \\
         \hline
         BERT & 3.20 $\pm$ 0.31 &7.49 $\pm$ 1.68 &21.92 $\pm$ 14.41 &58.94 $\pm$ 3.05 &62.30 $\pm$ 4.46 &10.85 $\pm$ 3.53 &9.06 $\pm$ 2.46 &17.04 $\pm$ 8.72 \\
         F1 & 6.40 $\pm$ 0.32 &16.07 $\pm$ 1.97 &45.79 $\pm$ 16.07 &69.01 $\pm$ 3.62 &73.72 $\pm$ 4.79 &23.39 $\pm$ 4.22 &19.53 $\pm$ 2.87 &35.39 $\pm$ 9.73\\
         BLEU  & 6.90 $\pm$ 0.39 &14.96 $\pm$ 1.77 &46.63 $\pm$ 16.93 &68.66 $\pm$ 3.50 &72.97 $\pm$ 4.50 &24.33 $\pm$ 4.64 &19.97 $\pm$ 3.01 &35.66 $\pm$ 9.95\\
         METEOR & 6.82 $\pm$ 0.38 &15.93 $\pm$ 1.93 &54.09 $\pm$ 8.47 &79.20 $\pm$ 0.02 &85.55 $\pm$ 0.01 &30.35 $\pm$ 1.29 &25.88 $\pm$ 0.51 &36.00 $\pm$ 9.70 \\
         \hline
         \rowcolor{LightBlue}\multicolumn{9}{c}{LSTM Seq2Seq } \\
         \hline
         BERT &9.16 $\pm$ 0.27 &18.10 $\pm$ 2.47 &60.55 $\pm$ 4.58 &77.91 $\pm$ 0.03 &84.43 $\pm$ 0.02 &34.68 $\pm$ 1.90 &27.23 $\pm$ 0.79 &45.12 $\pm$ 7.11\\
         F1  & 12.92 $\pm$ 0.01 &26.47 $\pm$ 0.04 &74.63 $\pm$ 0.03 &78.44 $\pm$ 0.00 &84.05 $\pm$ 0.01 &43.66 $\pm$ 0.01 &31.83 $\pm$ 0.01 &61.11 $\pm$ 0.01 \\
         BLEU  &12.76 $\pm$ 0.01 &26.94 $\pm$ 0.04 &75.03 $\pm$ 0.03 &78.16 $\pm$ 0.00 &83.90 $\pm$ 0.00 &43.92 $\pm$ 0.01 &31.96 $\pm$ 0.01 &61.13 $\pm$ 0.00\\
         METEOR & 12.97 $\pm$ 0.00 &25.97 $\pm$ 0.03 &74.37 $\pm$ 0.01 &78.42 $\pm$ 0.00 &84.03 $\pm$ 0.01 &43.79 $\pm$ 0.04 &31.63 $\pm$ 0.02 &61.22 $\pm$ 0.02 \\
         \hline
         \rowcolor{LightBlue}\multicolumn{9}{c}{Bi-LSTM Seq2Seq + Attention} \\
         \hline
         BERT & 12.83 $\pm$ 0.10 &23.74 $\pm$ 0.13 &71.48 $\pm$ 1.07 &78.54 $\pm$ 0.07 &85.60 $\pm$ 0.00 &35.96 $\pm$ 0.72 &26.88 $\pm$ 0.07 &50.57 $\pm$ 1.36  \\
         F1 & 14.92 $\pm$ 0.00 &26.67 $\pm$ 0.07 &78.01 $\pm$ 0.01 &81.02 $\pm$ 0.06 &86.17 $\pm$ 0.00 &40.61 $\pm$ 0.00 &29.38 $\pm$ 0.01 &57.91 $\pm$ 0.01 \\
         BLEU & 15.13 $\pm$ 0.01 &25.87 $\pm$ 0.05 &78.11 $\pm$ 0.02 &80.43 $\pm$ 0.02 &86.20 $\pm$ 0.00 &40.82 $\pm$ 0.01 &29.91 $\pm$ 0.02 &57.76 $\pm$ 0.00 \\
         METEOR & 14.81 $\pm$ 0.00 &26.53 $\pm$ 0.07 &78.04 $\pm$ 0.01 &80.02 $\pm$ 0.01 &86.25 $\pm$ 0.00 &41.02 $\pm$ 0.00 &30.11 $\pm$ 0.02 &57.90 $\pm$ 0.01\\
         \hline
         \rowcolor{LightBlue}\multicolumn{9}{c}{Transformer Architecture} \\
         \hline
         BERT  & 11.81 $\pm$ 0.04 &9.01 $\pm$ 0.06 &65.01 $\pm$ 0.09 &76.23 $\pm$ 0.02 &84.38 $\pm$ 0.01 &20.60 $\pm$ 0.00 &18.87 $\pm$ 0.02 &15.48 $\pm$ 0.14 \\
         F1  & 17.97 $\pm$ 0.64 &11.26 $\pm$ 0.17 &71.08 $\pm$ 0.24 &77.82 $\pm$ 0.03 &85.27 $\pm$ 0.01 &22.47 $\pm$ 0.02 &19.06 $\pm$ 0.03 &20.24 $\pm$ 0.34 \\
         BestBLEU  & 10.43 $\pm$ 0.14 &9.71 $\pm$ 0.00 &64.42 $\pm$ 0.88 &76.10 $\pm$ 0.07 &84.20 $\pm$ 0.01 &19.83 $\pm$ 0.00 &18.34 $\pm$ 0.03 &15.35 $\pm$ 0.54\\
         METEOR & 10.77 $\pm$ 0.37 &7.92 $\pm$ 0.11 &63.64 $\pm$ 0.80 &76.58 $\pm$ 0.05 &84.50 $\pm$ 0.01 &20.17 $\pm$ 0.06 &18.38 $\pm$ 0.01 &15.03 $\pm$ 0.72\\
         \hline
    \end{tabular}
    \caption{Comparison of models selected different selection metrics on probe tasks in MultiWoZ dialogue data set. The performance is measured with \emph{F1} on the probetasks.}
    \label{tab:multiwoz-probe-tasks-metric}
\end{table*}

\end{document}